\documentclass{article}

\usepackage[final]{neurips_2020} 

\usepackage[utf8]{inputenc} 
\usepackage[T1]{fontenc}    
\usepackage{hyperref}       
\usepackage{url}            
\usepackage{booktabs}       
\usepackage{amsfonts}       
\usepackage{nicefrac}       
\usepackage{microtype}      

\usepackage[english]{babel}
\usepackage{amsmath,amssymb,bm,amsthm,enumitem,bbold,subfig,xcolor}
\usepackage[mathscr]{eucal}
\usepackage[short,c1]{optidef} 

\newcommand{\br}{\mathbf{r}}
\newcommand{\BB}{\mathbb{B}}
\newcommand{\NN}{\mathbb{N}}
\newcommand{\RR}{\mathbb{R}}
\newcommand{\Fk}{\mathscr{F}_k}
\newcommand{\Psc}{\mathscr{P}}
\newcommand{\Lcal}{\mathcal{L}}
\newcommand{\Bsc}{\mathscr{B}}
\newcommand{\Csc}{\mathscr{C}}
\newcommand{\X}{\mathscr{X}}
\DeclareMathOperator*{\Sp}{span}

\newcommand{\F}{\mathscr{F}} 
\renewcommand{\d}{\mathrm{d}} 
\newcommand{\tb}{\textbf} 
\newcommand{\R}{\RR} 
\newcommand{\N}{\NN} 
\newcommand{\Np}{\N^*} 
\renewcommand{\b}{\mathbf}
\renewcommand{\P}{\mathbb{P}} 
\newcommand{\C}{\mathcal{C}} 
\newcommand{\p}{\partial}

\newtheorem*{mainthm}{Theorem}

\DeclareMathOperator*{\argmin}{arg\,min}

\title{Hard Shape-Constrained Kernel Machines}
\author{%
	Pierre-Cyril Aubin-Frankowski \\
	\'Ecole des Ponts ParisTech and CAS \\
	MINES ParisTech, PSL\\
	Paris, 75006, France\\
	\texttt{pierre-cyril.aubin@mines-paristech.fr} \\
	\And
	Zolt{\'a}n Szab{\'o} \\
	Center of Applied Mathematics, CNRS\\
	{\'E}cole Polytechnique, Institut Polytechnique de Paris\\
	Route de Saclay, Palaiseau, 91128, France\\
	\texttt{zoltan.szabo@polytechnique.edu}
}

\begin{document}
	
	\maketitle
	
	\begin{abstract}
		Shape constraints (such as non-negativity, monotonicity, convexity) play a central role in a large number of applications, as they usually improve performance for small sample size and help interpretability. However enforcing these shape requirements in a \emph{hard} fashion is an extremely challenging problem. Classically, this task is tackled (i) in a soft way (without out-of-sample guarantees), (ii) by specialized transformation of the variables on a case-by-case basis, or (iii) by using highly restricted function classes, such as polynomials or polynomial splines. In this paper, we prove that hard affine shape constraints on function derivatives can be encoded in kernel machines which represent one of the most flexible and powerful tools in machine learning and statistics. Particularly, we present a tightened second-order cone constrained reformulation, that can be readily implemented in convex solvers. We prove performance guarantees on the solution, and demonstrate the efficiency of the approach in joint quantile regression with applications to economics and to the analysis of aircraft trajectories, among others.
	\end{abstract}
	
	\section{Introduction}
	Shape constraints (such as non-negativity, monotonicity, convexity) are omnipresent in data science with numerous successful applications in statistics, economics, biology, finance, game theory, reinforcement learning and control problems.  For example, in biology, monotone regression techniques have been applied to identify genome interactions \citep{luss12efficient}, and in dose-response studies \citep{hu05analysis}. Economic theory dictates that utility functions are increasing and concave \citep{matzkin91semiparametric}, demand functions of normal goods are downward sloping \citep{lewbel10shapeinvariant,blundell12measuring}, production functions are concave \citep{varian84nonparametric} or S-shaped \citep{yagi20shapeconstrained}. Moreover cyclic monotonicity has recently turned out to be beneficial in panel multinomial choice problems \citep{shi18estimating}, and most link functions used in a single index model are monotone \citep{li07nonparametric,chen16generalized,balabdaoui19least}. Meanwhile, supermodularity is a common assumption in supply chain models,  stochastic multi-period inventory problems, pricing models and game theory \citep{topkis98supermodularity,simchilevi14logic}. In finance, European and American call option prices are convex and monotone in the underlying stock price and increasing in volatility \citep{aitsahalia03nonparametric}. 
	In statistics,  the conditional quantile function is increasing w.r.t.\ the quantile level.  
	In reinforcement learning and in stochastic optimization the value functions are regularly supposed to be convex \citep{keshavarz11imputing,shapiro14lectures}.  More examples can be found in recent surveys on shape-constrained regression \citep{johnson18shape,guntuboyina18nonparametric,chetverikov2018econometrics}.

	Leveraging prior knowledge expressed in terms of shape structures has several practical benefits: the resulting techniques allow for estimation with smaller sample size and 
	the imposed shape constraints help interpretability. Despite the numerous practical advantages, the construction of shape-constrained estimators can be quite challenging. Existing techniques typically impose the shape constraints (i) in a 'soft' fashion as a regularizer or at finite many points \citep{delecroix96functional,blundell12measuring,aybat14parallel,wu15penalized,takeuchi06nonparametric,sangnier16joint,chen16generalized,agrell19gaussian,mazumder19computational,koppel19projected,han19isotonic,yagi20shapeconstrained}, (ii) through constraint-specific transformations of the variables such as quadratic reparameterization \citep{flaxman17poisson}, positive semi-definite quadratic forms \citep{bagnell15learning}, or integrated exponential functions \citep{wu18semiparametric}, or (iii) they make use of highly restricted functions classes such as classical polynomials \citep{hall2018thesis} or polynomial splines \citep{turlach05shape,papp14shape,pya15shape,wu18semiparametric,meyer18framework,koppel19projected}.  Both the polynomial and spline-based shape-constrained techniques rely heavily on the underlying algebraic structure of these bases, without direct extension to more general function families. 
	
	From a statistical viewpoint, the main focus has been on the design of estimators with uniform guarantees \citep{horowitz17nonparametric,freyberger18inference}. Several existing methods have been analyzed from this perspective and were shown to be (uniformly) consistent, on a case-by-case basis and when handling specific shape constraints \citep{wu15penalized,chen16generalized,han16multivariate,mazumder19computational,koppel19projected,han19isotonic,yagi20shapeconstrained}. While these asymptotic results are of significant theoretical interest, applying shape priors is generally beneficial in the small sample regime. 
	In this paper we propose a flexible \tb{optimization framework} allowing multiple shape constraints to be jointly handled in a hard fashion. In addition, to address the bottlenecks of restricted shape priors and function families, we consider general affine constraints on derivatives, and use reproducing kernel Hilbert spaces (RKHS) as hypothesis space.
	
	RKHSs (also called abstract splines; \citealp{aronszajn50theory,wahba90spline,berlinet04reproducing,wang11splines}) increase significantly the richness and modelling power of classical polynomial splines. Indeed, the resulting function family can be rich enough for instance (i) to encode probability distributions without loss of information \citep{fukumizu08kernel,sriperumbudur10hilbert}, (ii) to characterize statistical independence of random variables \citep{bach02kernel,szabo18characteristic}, or (iii) to approximate various function classes arbitrarily well \citep{steinwart01influence,micchelli06universal,carmeli10vector,sriperumbudur11universality,simon-gabriel18kernel}, including the space of bounded continuous functions.
	An excellent overview on kernels and RKHSs is given by \citet{hofmann08kernel,steinwart08support,saitoh16theory}.
	
	In this paper we incorporate into this flexible RKHS function class the possibility to	impose \emph{hard} linear shape requirements on derivatives, i.e.\ constraints of the form
	\begin{align}
	0 & \le b+Df(\b x)   \quad \forall \,\b x \in K \label{eq:pointwise-constraints}
	\end{align}
	for a bias $b\in\R$, given a differential operator ${D=\sum_{j}\gamma_j \p^{\br_j}}$ where $\p^{\b r}f(\b x)=\frac{\p^{\sum_{j=1}^dr_j}f(\b x)}{\p^{r_1}_{x_1}\cdots\p^{r_d}_{x_d}}$ and a  compact set $K\subset \R^d$. The fundamental technical challenge is to guarantee the pointwise inequality \eqref{eq:pointwise-constraints} at \emph{all} points of the (typically non-finite) set $K$. We show that one can tighten the infinite number of affine constraints \eqref{eq:pointwise-constraints} into a finite number of second-order cone constraints
	\begin{align}
	\eta \|f\| & \le b+ Df(\b x_m)   \quad \forall \, m \in  \{1,\ldots,M\} \label{eq:pointwise-constraints_SOC}
	\end{align}
	for a suitable choice of $\eta>0$ and $\{\b x_m\}_{m=1\dots M} \subseteq K$.
	
	Our \tb{contributions} can be summarized as follows.
	\begin{enumerate}[labelindent=0cm,leftmargin=*,topsep=0cm,partopsep=0cm,parsep=0.1cm,itemsep=0cm]
		\item We show that hard shape requirements can be embedded in kernel machines by taking  a second-order cone (SOC) tightening of constraint \eqref{eq:pointwise-constraints}, which can be readily tackled by convex solvers. Our formulation builds upon the reproducing property for kernel derivatives and on coverings of compact sets. 
		\item We prove error bounds on the distance between the solutions of the strengthened and original problems.
		\item  We achieve state-of-the-art performance in joint quantile regression (JQR) in RKHSs. We also combine JQR with other shape constraints in economics and in the analysis of aircraft trajectories.
	\end{enumerate}	
	The paper is structured as follows.  Section~\ref{sec:problem-formulation} is about problem formulation. Our main result is presented in 
	Section~\ref{sec:result}. Numerical illustrations are given in Section~\ref{sec:numerical-demo}.
	Proofs and additional examples are provided in the supplement.
	
	\section{Problem formulation} \label{sec:problem-formulation}
	In this section we formulate our problem after introducing some notations, which the reader may skip at first, and return to if necessary.
	
	\textbf{Notations:} Let $\NN:= \{0,1,\dots\}$, $\Np:= \{1,2,\dots\}$ and $\RR_+$ denote the set of natural numbers, positive integers and non-negative real numbers, respectively. We use the shorthand $[n]:=\{1,\dots,n\}$.
	The $p$-norm of a vector $\b v\in\R^p$ is $\left\|\b v \right\|_p = (\sum_{j\in[d]}|v_j|^p)^{\frac{1}{p}}$ ($1\le p <\infty$); 
	$\left\|\b v\right\|_{\infty} = \max_{j\in [d]}|v_j|$.	The $j$-th canonical basis vector is $\b e_j$; $\b 0_d\in \R^d$ is the zero vector. 
	Let $\BB_{\|\cdot\|}(\b c,r)=\{\b x \in \R^d: \left\|\b x - \b c\right\|\le r\}$ be the closed ball in $\R^d$ with center $\b c$ and radius $r$ in norm $\left\|\cdot\right\|$. Given a norm $\left\|\cdot\right\|$ and radius $\delta>0$, a  
	$\delta$-net of a compact set $K\subset \R^d$ consists of a set of points $\{\b x_m\}_{m\in[M]}$ such that $K\subseteq \cup_{m\in[M]} \BB_{\|\cdot\|}(\b x_m,\delta)$, in other words $\left\{\BB_{\|\cdot\|}(\b x_m,\delta)\right\}_{m\in[M]}$ forms a covering of $K$. The identity matrix is $\b{I}$. For a matrix $\b M\in \R^{d_1\times d_2}$, $\b M^{\top} \in \R^{d_2\times d_1}$ denotes its transpose, 
	its operator norm is $\left\|\b M\right\|=\sup_{\b x \in \R^{d_2}: \|\b{x}\|_2 = 1}\left\|\b M\b x\right\|_2$. The inverse of a non-singular matrix $\b M\in\R^{d\times d}$ is $\b M^{-1} \in \R^{d \times d}$. The concatenation of  matrices $\b M_1\in\R^{d_1 \times d}, \ldots, \b M_N\in\R^{d_N \times d}$ is denoted by $\b M=\left[\b M_1;\ldots;\b M_N\right]\in \R^{\left(\sum_{n\in[N]}d_n\right) \times d}$.  Let $\X$ be an open subset of $\R^d$ with a real-valued kernel $k:\X \times \X \rightarrow \R$, and associated reproducing kernel Hilbert space (RKHS) $\Fk$. The Hilbert space $\Fk$ is characterized by $f(\b x) = \left<f,k( \b x, \cdot)\right>_{k}$ ($\forall \b x \in \X, \forall f \in \Fk$) and $k(\b x, \cdot) \in \Fk$ ($\forall  \b x \in \X$) where $\left<\cdot,\cdot\right>_{k}$ stands for the inner product in $\Fk$, and $k(\b x,\cdot)$ denotes the function $\b y \in \X \mapsto k(\b x,\b y) \in \R$. The first property is called the reproducing property, the second one describes a generating family of $\Fk$. The norm on $\Fk$ is written as $\left\|\cdot\right\|_k$.  
	For a multi-index $\b r\in \N^d$ let  the $\b r$-th order partial derivative of a function $f$ be denoted by $\p^{\b r}f(\b x)=\frac{\p^{|\b r|}f(\b x)}{\p^{r_1}_{x_1}\cdots\p^{r_d}_{x_d}}$ where $|\br| = \sum_{j\in[d]} r_j$ is the length of $\br$. When $d=1$ the $f^{(n)}=\p^{n}f$ notation is applied; specifically $f''$ and $f'$ are used in case of $n=2$ and $n=1$.
	Given $s\in \N$, let $\C^s(\X)$ be the set of real-valued functions on $\X$ with continuous derivatives up to order $s$ (i.e., $\p^{\br}f \in \C(\X):=\C^0(\X)$ when $|\br |\le s$). 
	Let $I \in \Np$. Given $(A_i)_{i\in[I]}$ sets let $\prod_{i\in [I]}A_i$ denote their 
	Cartesian product; we write $A^I$ in case of $A=A_1=\ldots=A_{I}$. 
	
	Our \tb{goal} is to solve hard shape-constrained kernel machines of the form
	\begin{align}
	\left(\bar{\b f},\bar{\bm{b}}\right) &= \argmin_{\b f=(f_q)_{q\in [Q]}\,\in\,(\Fk)^Q,\, \b{b}=(b_p)_{p\in[P]}\,\in\,\Bsc,\,(\b f, \b b)\,\in \, C
	} \hspace*{-2cm}\Lcal(\b f,\bm{b}),
	\label{opt-cons}\tag{$\Psc$}
	\end{align}
	where we are given an objective function $\Lcal$ and a constraint set $C$ (detailed below),
	a closed convex constraint set $\Bsc\subseteq \RR^{P}$ on the biases,   an order $s\in \NN$,
	an open set $\X \subseteq \R^d$ with a kernel $k\in \mathcal{C}^{s}(\X\times \X)$ and associated RKHS $\F_k$, and samples $S=\left\{(\b x_n,y_n)\right\}_{n\in[N]} \subset \X \times \R$. The objective function in \eqref{opt-cons} is specified by the triplet $(S,L,\Omega)$:
	\begin{align*}
	\Lcal(\b f,\bm{b}) & = L\left(\b{b},\left(\b x_n,y_n,\left(f_q(\b x_n)\right)_{q\in[Q]}\right)_{n\in [N]}\right) + \Omega\left(\left(\|f_q\|_k\right)_{q\in [Q]}\right),
	\end{align*}
	with loss function $L: \R^P \times \left(\X \times \R\times \R^Q\right)^N \rightarrow \R$ and regularizer $\Omega:(\RR_+)^Q \rightarrow \R$. Notice that the objective $\Lcal$ depends on the samples $S$ which are assumed to be fixed, hence our proposed optimization framework focuses on the empirical risk.
	The bias $\b{b}\in\R^{P}$ can be both constraint (such as $f(x)\ge b_1$, $f'(x)\ge b_2$) and variable-related ($f_q+b_q$, see \eqref{def_QR_loss}-\eqref{def_QR_cons} later). The restriction of $L$ to $\Bsc$ is assumed to be strictly convex in $\b b$, and $\Omega$ is supposed to be strictly increasing in each of its arguments to ensure the uniqueness of minimizers of \eqref{opt-cons}.  
	
	The $I\in\Np$ hard shape requirements in \eqref{opt-cons} take the form\footnote{In constraint  \eqref{def_mixded_constraint}, $\b W \b f$ is meant as a formal matrix-vector product: $(\b W \b f)_i = \sum_{q\in[Q]}W_{i,q}f_q$.}
	\begin{align}
	C & =\left\{(\b f,\b b)\,|\, (\b b_0 - \b U\b b)_i \le  D_i  (\b W \b f - \b f_0)_i(\b x), \forall \b x \in K_i, \forall i\in [I]\right\},
	\label{def_mixded_constraint}\tag{$\Csc$}
	\end{align}
	i.e., \eqref{def_mixded_constraint} encodes affine constraints of at most $s$-order derivatives ($D_i = \sum_{j\in [n_{i,j}]}\gamma_{i,j}\p^{\b r_{i,j}},\, |\b r_{i,j}|\le s$, $\gamma_{i,j}\in \R$). Possible shifts are expressed by the terms $\b b_0 = (b_{0,i})_{i\in [I]}\in\RR^I$,  $\b f_0 = (f_{0,i})_{i\in [I]}\in \left(\Fk\right)^I$. The matrices $\b U \in \R^{I\times P}$ and $\b W \in \R^{I\times Q}$ capture the potential interactions within the 
	bias variables $(b_p)_{p\in[P]}$ and functions $(f_q)_{q\in[Q]}$, respectively. The 
	compact sets $K_i\subset \X$ ($i\in [I]$) define the domain where the constraints are imposed.   

	\noindent\tb{Remarks:} 
	\begin{itemize}[labelindent=0cm,leftmargin=*,topsep=0cm,partopsep=0cm,parsep=0.1cm,itemsep=0cm]
		\item Differential operators: As 
		$\X \subseteq \R^d$ is open and $k\in \mathcal{C}^{s}(\X\times \X)$, any differential operator $D_i$ of order at most $s$ is well defined \citep[Theorems 2.5 and 2.6, page~76]{saitoh16theory} as a 
		mapping from $\Fk$ to $\mathcal{C}(\X)$. Since the coefficients $\{\gamma_{i,j}\}_{j\in[n_{i,j}]}$ of $D_i$-s belong to the whole $\RR$, \eqref{def_mixded_constraint} can cover inequality constraints in both directions.
		\item Bias constraint $\Bsc$: Choosing $\Bsc=\{\b{0}_P\}$ leads to constant l.h.s. $\b b_0$ in \eqref{def_mixded_constraint}. The other extreme is 
		$\Bsc=\RR^P$ in which case no hard constraint is imposed on the bias variable $\b b$.
		\item Compactness of $K_i$-s: The compactness assumption on the sets $K_i$ is exploited in the construction of our strengthened optimization problem (Section~\ref{sec:result}). 
		This requirement also ensures not imposing
		restrictions ''too far'' from the observation points, which could be impossible to satisfy. Indeed, let us consider for instance a $c_0$-kernel $k$ on $\R$, i.e.\ that $k(x,\cdot)\in \C^0(\RR)$ for all $x$ and $\lim_{|y|\rightarrow \infty}k(x,y) = 0$ for all $x \in \R$ (as for the Gaussian kernel). In this case $\lim_{|y|\rightarrow \infty} f(y) = 0$ also holds for all $f\in \Fk$.
		Hence a constraint of the form ``for all $t\in\RR_+, f(t) \geq \epsilon > 0$`` can \emph{not} be satisfied for $f \in \Fk$.
		\item Assumption on $\X$: 
		If $s=0$ (in other words only function evaluations are present in the shape constraints), then $\X$ can be any topological space. 
	\end{itemize}
	
	We give various \tb{examples} for the considered problem family \eqref{opt-cons}.  We start with an example where $Q=1$. 
	
	\noindent\tb{Kernel ridge regression} (KRR) with \emph{monotonicity} constraint: In this case the objective function and constraint are  
	\begin{align}
	\Lcal(f,b) &:=\frac{1}{N} \sum_{n \in [N]} |y_n-f(x_n)|^2+\lambda_f \|f\|^2_k, \text{ s.t. }
	f'(x) \geq 0,\, \forall x\in [x_l,x_u] \label{def_KRR}
	\end{align}
	with $\lambda_f>0$. In other words in \eqref{opt-cons} we have $Q=1$, $d=1$, $s=1$, $P=I=1$, $K_1=[x_l,x_u]$, $\Omega(z)=\lambda_f z^2$, $D_1 = \partial^1$, $U=W=1$, $f_{1,0} = 0$, $b_{1,0} = 0$, and $b\in \Bsc=\{0\}$ is a dummy variable. 
	
	\noindent \tb{Joint quantile regression} (JQR; e.g. \citealp{sangnier16joint}): Given $0<\tau_1<\ldots<\tau_Q<1$ levels the goal is to estimate \emph{jointly} the $(\tau_1,\dots,\tau_Q)$-quantiles of the conditional
	distribution $\P(Y|X= \b x)$ where $Y$ is real-valued. In this case the objective function is 
	\begin{align}\label{def_QR_loss} 
	\Lcal\left(\b f,\b{b}\right) &=\frac{1}{N}\sum_{q\in [Q]}\sum_{n \in [N]} l_{\tau_q}\left(y_n-[f_q(\b x_n)+b_q]\right)   + \lambda_{\b{b}} \|\b{b}\|^2_2 + \lambda_f \sum_{q \in [Q]}\|f_q\|^2_k, 
	\end{align}
	where $\lambda_{\b{b}}> 0$, $\lambda_f>0$,\footnote{\citet{sangnier16joint} considered the same loss but a \emph{soft} non-crossing inducing regularizer inspired by matrix-valued kernels, and also set $\lambda_{\b{b}}= 0$.} and the pinball loss is defined as $l_\tau(e) = \max(\tau e , (\tau-1)e)$  with $\tau \in (0,1)$. In JQR, the estimated $\tau_q$-quantile functions $\{f_q+b_q\}_{q\in[Q]}$ are \emph{not} independent; the joint shape constraint they should satisfy is the monotonically increasing property w.r.t.\ the quantile level $\tau$. It is natural to impose this \emph{non-crossing} requirement on the smallest rectangle containing the points $\{\b x_n\}_{n\in [N]}$, i.e. $K~=~\prod_{r\in [d]}\left[\min \{( \b x_n)_r\}_{n\in [N]},\max \{(\b x_n)_r\}_{n\in [N]}\right]$. The corresponding shape constraint is
	\begin{equation}\label{def_QR_cons}
	f_{q+1}(\b x)+b_{q+1}\geq f_{q}(\b x)+b_{q},\, \forall q\in [Q-1],\, \forall  \b x\in K.
	\end{equation} 	
	One gets \eqref{def_QR_loss}-\eqref{def_QR_cons} from \eqref{opt-cons}  by choosing
	$P=Q$, $I=Q-1$, $s=0$, $\b b_0 =\b 0$, 
	$\b f_0 = \b 0$, $\Bsc = \R^P$, 
	$K_i = K$ ($\forall i \in [I]$), $\Omega(\b z)=\lambda_f \sum_{q\in [Q]}(z_q)^2$,  and 
	$\b U = \b W =
	\begin{bmatrix}
	-1 & 1 & 0 & 0\\
	0 & -1 & 1 & 0\\
	\vdots & \hspace{-.7cm}\ddots & \hspace{-.7cm}\ddots & \hspace{-.7cm}\ddots  \\
	0 & 0 & -1 & 1
	\end{bmatrix}        \in \R^{(Q-1)\times Q}$.\\
	\noindent \tb{Further examples}: There are various other widely-used shape constraints beyond non-negativity (for which $s=0$), monotonicity ($s=1$) or convexity ($s=2$) which can be taken into account in \eqref{def_mixded_constraint}. For instance one can consider
	$n$-monotonicity ($s=n$), 
	$(n-1)$-alternating monotonicity, 
	monotonicity w.r.t.\ unordered weak majorization ($s=1$) 
	or w.r.t.\ product ordering ($s=1$), or supermodularity ($s=2$). For details on how these shape constraints can be written as \eqref{def_mixded_constraint}, see the supplement (Section~\ref{sec:shape-constraints}).
	
	\section{Results} \label{sec:result}
	In this section, we first present our strengthened SOC-constrained problem, followed by a representer theorem and explicit bounds on the distance to the solution of \eqref{opt-cons}.
	
	In order to introduce our proposed tightening, let us first consider the discretization of the $I$ constraints using $M_i$ points $\left\{\tilde{\b x}_{i,m}\right\}_{m\in[M_i]}\subseteq K_i$. This would lead to the following relaxation of \eqref{opt-cons}
	\begin{align}
	v_{\text{disc}}  & = \min_{\b f\,\in\,(\Fk)^Q,\,\bm{b}\,\in\,\Bsc} \Lcal(\b f,\bm{b})\, \text{ s.t. } 
	(\b b_0-\b U \b b)_i  \;\leq \min_{m\in[M_i]} D_i(\b W \b f-\b f_{0})_i\left(\tilde{\b x}_{i,m}\right) \,\forall i\in [I].
	\label{eq:v-disc}
	\end{align}
	Our proposed SOC-constrained tightening can be thought of as adding extra, appropriately chosen, $\eta_i$-buffers to the l.h.s. of the constraints:
	\begin{argmini!}
		{\substack{\b f\,\in\,(\Fk)^Q,\,\bm{b}\,\in\,\Bsc\subset \R^p}}{\hspace*{-0.2cm} \Lcal(\b f,\bm{b})}{}{\left(\b f_{\bm{\eta}},\b b_{\bm{\eta}}\right)=\label{opt-soc_inf}\tag{$\Psc_{\bm{\eta}}$}} 
		\addConstraint{\let\scriptstyle\textstyle \hspace{-1cm}(\b b_0-\b U \b b)_i + \eta_i \|(\b W \b f-\b f_{0})_i\|_k\, \;\leq \min_{m\in[M_i]} D_i(\b W \b f-\b f_{0})_i\left(\tilde{\b x}_{i,m}\right)}{\,\forall i\in [I]}{\label{def_soc_inf}\tag{$\Csc_{\bm{\eta}}$}}.
	\end{argmini!}	
	The SOC constraint \eqref{def_soc_inf} is determined by a fixed $\bm \eta = (\eta_i)_{i\in [I]}\in \R_+^{I}$ coefficient vector and by the points $\{\tilde{\b x}_{i,m}\}$.\footnote{Constraint \eqref{def_soc_inf} is the precise meaning of the preliminary intuition given in \eqref{eq:pointwise-constraints_SOC}.} For each $i\in [I]$, the  points $\left\{\tilde{\b x}_{i,m}\right\}_{m\in[M_i]}$ are chosen to form a $\delta_i$-net of the compact set $K_i$ for some $\delta_i>0$ and a pre-specified norm  $\left\|\cdot\right\|_\X$.\footnote{The existence of finite $\delta_i$-nets ($M_i<\infty$) stems from the compactness of $K_i$-s. The flexibility in the choice of the norm $\left\|\cdot\right\|_{\X}$ allows for instance using  
		cubes by taking the $\left\|\cdot\right\|_1$ or the $\left\|\cdot\right\|_\infty$-norm on $\RR^d$ when covering the $K_i$-s.} Given $\left\{\tilde{\b x}_{i,m}\right\}_{m\in[M_i]}$, the coefficients $\eta_i \in \R_+$ are then defined as 
	\begin{equation}\label{def_eta}
	\eta_i = \hspace{-0.5cm}\sup_{m\,\in\,[M_i],\, \b u \,\in\,\BB_{\|\cdot\|_\X}(\b 0,1)} \hspace*{-0.4cm} \|D_{i,\b x}k(\tilde{\b x}_{i,m},\cdot)-D_{i,\b x}k(\tilde{\b x}_{i,m}+\delta_i \b u,\cdot)\|_k,
	\end{equation} 
	where $D_{i,\b x}k(\b x_0,\cdot)$ is a shorthand for $\b y \mapsto D_i(\b x \mapsto k(\b x,\b y))(\b x_0)$.
	Problem \eqref{opt-soc_inf} has $I$ scalar SOC constraints \eqref{def_soc_inf} over infinite-dimensional variables. Let $\bar{v}=\Lcal\left(\bar{\b f},\bar{\bm{b}}\right)$ be the minimal value of \eqref{opt-cons} and $v_{\bm{\eta}}=\Lcal\left(\b f_{\bm{\eta}},\b b_{\bm{\eta}}\right)$ be that of \eqref{opt-soc_inf}. 
	Notice that, when formally setting $\bm{\eta}=\b 0$, \eqref{opt-soc_inf} corresponds to \eqref{eq:v-disc}.
	
	In our main result below (i) shows that \eqref{def_soc_inf} is indeed a tightening of \eqref{def_mixded_constraint}, (ii) provides a representer theorem which allows to solve numerically \eqref{opt-soc_inf}, and (iii) gives bounds on the difference between the solution of 
	\eqref{opt-soc_inf} and that of \eqref{opt-cons} as a function of $(v_{\bm \eta}-v_{\text{disc}})$ and $\bm{\eta}$ respectively.
	\begin{mainthm}[{Tightened task, representer theorem, bounds}]\label{thm_SOC_bounds} Let $\b f_{\bm{\eta}}=(f_{\bm{\eta},q})_{q\in[Q]}$.
		Then, 
		
		(i) Tightening: any $(\b f,\bm{b})$ satisfying \eqref{def_soc_inf} also satisfies \eqref{def_mixded_constraint}, hence $v_{\text{disc}}\leq \bar{v} \leq v_{\bm{\eta}}$.
		
		(ii) Representer theorem: For all $q\in [Q]$, there exist real coefficients $\tilde{a}_{i,0,q}, \tilde{a}_{i,m,q}, a_{n,q} \in \R$  such that
		$f_{\bm{\eta},q}=\sum_{i \in [I]}\left[\tilde{a}_{i,0,q}f_{0,i}+ \sum_{m\in[M_i]} \tilde{a}_{i,m,q} D_{i,\b x}k\left(\tilde{\b x}_{i,m},\cdot\right)\right] +\sum_{n\in[N]} a_{n,q}k(\b x_{n},\cdot)$.
		
		(iii) Performance guarantee: if 
		$\Lcal$ is $(\mu_{f_q},\mu_{\bm{b}})$-strongly convex w.r.t.\ $(f_q,\b{b})$ for any $q\in[Q]$, then
		\begin{align}\label{ineq_aposteriori}
		\|f_{\bm{\eta},q}-\bar{f}_q\|_k &\leq \sqrt{\frac{2 (v_{\bm{\eta}}-v_{\text{disc}})}{\mu_{f_q}}}, &
		\|\bm{\bm{b_\eta}}-\bar{\bm{b}}\|_2 &\leq \sqrt{\frac{2 (v_{\bm{\eta}}-v_{\text{disc}})}{\mu_{\bm{b}}}}.
		\end{align}
		If in addition  $\b U$ is of full row-rank (i.e. surjective), $\Bsc=\RR^P$, and $\Lcal( \bar{\b f},\cdot)$ is $L_b-$Lipschitz continuous on 
		$\BB_{\left\|\cdot\right\|_2}\left(\bar{\b b},c_f\|\bm{\eta}\|_\infty\right)$ where 
		$c_f~=~\sqrt{I}\left\|\left(\b U^{\top} \b U\right)^{-1}\b U^{\top}\right\| \max_{i\in [I]}\left\|(\b W \bar{\b f}-\b f_{0})_i\right\|_k$, then
		\begin{align}\label{ineq_apriori}
		\|f_{\bm{\eta},q}-\bar{f}_q\|_k &\leq \sqrt{\frac{2 L_b c_f \|\bm{\eta}\|_\infty}{\mu_{f_q}}}, &
		\|\bm{b_\eta}-\bar{\bm{b}}\|_2&\leq \sqrt{\frac{2 L_b c_f \|\bm{\eta}\|_\infty}{\mu_{\bm{b}}}}.
		\end{align}
	\end{mainthm}
	\tb{Proof} (idea): The SOC-based reformulation relies on rewriting the constraint \eqref{def_mixded_constraint} as the inclusion of the sets $\Phi_{D_i}(K_i)$ in the closed halfspaces $H^+_{\phi_i,\beta_i}:=\{g\in \Fk\,|\, \left\langle \phi_i,g \right\rangle_k\geq \beta_i\}$ for $\forall i\in [I]$ where $\Phi_{D_i}(\b x):= D_{i,\b x} k(\b x,\cdot)\in \Fk$, $\Phi_{D_i}(X):= \{\Phi_{D_i}(\b x)\,|\, \b x\in X\}$, $\phi_i:=(\b W \b f-\b f_{0})_i$ and $\beta_i:=(\b b_0-\b U \b b)_i$. The tightening is obtained by guaranteeing these inclusions with an $\eta_i$-net of 
	$\Phi_{D_i}(K_i)$ containing the $\delta_i$-net of $K_i$ when pushed to $\F_k$. The bounds stem from classical inequalities for strongly convex objective functions.  The proof details of (i)-(iii) are available in the supplement (Section~\ref{sec:proofs}).
	
	\noindent\tb{Remarks}: 
	
	The \tb{representer theorem} allows one to express \eqref{opt-soc_inf} as a finite-dimensional SOC-constrained problem:
	\begin{align*}
	\min_{\substack{\b A\,\in\,\RR^{N\times Q},\,\bm{b}\,\in\,\Bsc,\\ \tilde{\b A}\,\in\,\RR^{N\times Q},\,\tilde{\b A}_0\,\in\,\RR^{I\times Q}}} \hspace{-1cm}\Lcal(\b f,\bm{b})& \text{ s.t. } 
	(\b b_0-\b U \b b)_i + \eta_i \left\|\b{G}^{1/2} \b g_i \right\|_2 \hspace{-0.09cm} \leq \hspace{-0.15cm}\min_{m\in[M_i]} \left(\b{G}_{D_i} \b g_i\right)_{I+N+m}\, \forall i\in [I],\,\forall q\in [Q],     
	\end{align*}
	where $\tilde{\b e}_i \in \RR^{I}$ and $\b e_i \in \RR^{I+N+M}$ are the canonical basis vectors, $\b g_i := \left[\tilde{\b A}_0;\b A;\tilde{\b A}\right]\b W^{\top} \tilde{\b e}_i - \b e_i$ and the coefficients of the components of $\b f$
	were collected to
	$\tilde{\b A}_0 = [\tilde{a}_{i,0,q}]_{i\in [I],\, q\in [Q]}\in\R^{I\times Q}$, 
	$\b A = \left[a_{n,q}\right]_{n\in [N],\, q\in [Q]} \in \R^{N\times Q}$,
	$\tilde{\b A}=\left[\tilde{\b a}_{i,q}\right]_{i\in[I],\, q\in [Q]}\in\R^{M\times Q}$ 
	with $M=\sum_{i \in [I]} M_i$ and $\tilde{\b a}_{i,q}=\left[\tilde{a}_{i,m,q}\right]_{m\in [M_i]}\in\R^{M_i}$ ($i\in[I]$, $q\in [Q]$).	
	In this finite-dimensional optimization task, 
	$\b{G}\in\R^{(I+N+M)\times (I+N+M)}$ is the Gram matrix of $(\{f_{0,i}\}_{i\in{I}},\{k(\b x_{n},\cdot)\}_{n\in[N]},\{D_{i,\b x}k(\tilde{\b x}_{i,m},\cdot)\}_{m\in[M_i], i\in{I}})$, $\b{G}_{D_i}\in\R^{(I+N+M)\times (I+N+M)}$ is the Gram matrix of the differentials $D_i$ of these functions, 
	$\b{G}^{1/2}$ is the matrix square root of the positive semi-definite $\b{G}$.
	
	The \tb{bounds}\footnote{Notice that \eqref{ineq_aposteriori} is a computable bound, while \eqref{ineq_apriori} is not, as the latter depends on properties of the unknown solution of \eqref{opt-cons}.} \eqref{ineq_aposteriori}-\eqref{ineq_apriori} show that  smaller $\bm{\eta}$ gives tighter guarantee on the recovery of $\bar{\b f}$ and $\bar{\b b}$.  
	Since $\left|\p^{\b{r}}f_{\bm{\eta},q}(\b x)-\p^{\b{r}}\bar{f}_q(\b x)\right| \leq \sqrt{\p^{\b{r},\b{r}}k(\b x,\b x)}\left\|f_{\bm{\eta},q}-\bar{f_q}\right\|_k$ by the reproducing property and the Cauchy-Schwarz inequality, the bounds on 
	$\|f_{\bm{\eta}}-\bar{f}\|_k$ can be propagated to pointwise bounds on the derivatives (for $|\br|\le s$). We emphasize again that in our optimization problem \eqref{opt-soc_inf} the samples $S=\{(\b x_n,y_n)\}_{n\in [N]}$ are assumed to be fixed; in other words the bounds \eqref{ineq_aposteriori} and \eqref{ineq_apriori} are meant conditioned on $S$.
	
	The \tb{parameters} $M$, $\b \delta$ and $\bm \eta$ are strongly intertwined, their interplay reveals  an \textit{accuracy-computation tradeoff}. Consider a shift-invariant kernel ($k(\b x,\b y)=k_0(\b x-\b y)$, $\forall \b x,\b y$), then \eqref{def_eta} simplifies to 
	$\eta_i :=\sup_{\b u\in \BB_{\|\cdot\|_\X}(\b 0,1)}\hspace*{-0.1cm} \sqrt{ \left|2 D_{i,\b x}D_{i,\b y}k_0(\b 0)- 2D_{i,\b x}D_{i,\b y}k_0\left(\delta_i \b u\right)\right|}$, where $D_{i,\b y}$ is defined similarly to $D_{i,\b x}$.\footnote{Similar computation can be carried out for higher order derivatives. For more general kernels, estimating $\eta_i$-s can be
    also done by sampling uniformly $\b u$ in the unit ball.} 
	This expression of $\eta_i$ implies that whenever $D_{i,\b x}D_{i,\b y}k_0$ is $L_\delta$-Lipschitz\footnote{For instance any $C^{s+1}$ kernel satisfies this local Lipschitz requirement.} on $\BB_{\|\cdot\|_\X}(\b 0,\delta_i)$, then $\eta_i\leq \sqrt{2L_\delta} \sqrt{\delta_i}$. By the previous point, a smaller $\bm{\eta}$ ensures a better recovery which can be guaranteed by smaller $\delta_i$-s, which themselves correspond to a larger number of centers ($M_i$-s) in the $\delta_i$-nets of the $K_i$-s. Hence, one can control the computational complexity by the total number $M$ of points in the nets.  Indeed, most SOCP solvers 
	rely on primal-dual interior point methods which have (in the worst-case) cubic complexity $\mathcal{O}\left((P+N+M)^3\right)$ per iterations \citep{alizadeh03second}. Controlling $M$ allows one to tackle hard shape-constrained problems in moderate dimensions ($d$); for details see Section~\ref{sec:numerical-demo}. In practice, to reduce the number of coefficients in $f_{\eta,q}$, it is beneficial to recycle 
	the points $\{\b x_n\}_{n\in[N]}$ among the $M_i$ virtual centers, whenever the points belong to a constraint set $K_i$. This simple trick was possible in all our numerical examples and kept the computational expense quite benign. Supplement (Section~\ref{sec:KRR-demo}) presents an example of the actual computational complexity observed.
	
	While in this work we focused on the optimization problem \eqref{opt-cons} which contains \emph{solely} infinite-dimensional SOC constraints \eqref{def_soc_inf}, the proved  \eqref{def_soc_inf} $\Rightarrow$ \eqref{def_mixded_constraint} implication can be of independent interest to tackle problems where other types of constraints are present.\footnote{For example having a unit integral is a natural additional requirement beyond non-negativity in density estimation, and writes as a linear equality constraint over the coefficients of $f_{\bm{\eta},q}$.} For simplicity we formulated our result with uniform coverings ($\delta_i$, $\eta_i$, $i\in [I]$). However, we prove it 
	for more general non-uniform coverings ($\delta_{i,m}$, $\eta_{i,m}$, $i\in [I]$, $m\in[M_i]$; see Section~\ref{sec:proofs}). This can beneficial for sets with complex geometry (e.g. star-shaped) or when recyling of the samples was used to obtain coverings (as the samples in $S$ have no reason to be equally spaced); we provide an example (in economics) using a non-uniform covering in Section~\ref{sec:numerical-demo}.
	
	In practice, since the convergence speed of SOCP solvers depends highly on the condition number of $\b{G}^{1/2}$, it is worth replacing $\b{G}^{1/2}$ with $(\b{G}+\epsilon_{\text{tol}}\b{I})^{1/2}$, setting a tolerance $\epsilon_{\text{tol}}\simeq 10^{-4}$. As $\bm{G}+\epsilon_{\text{tol}}\b{I}\succcurlyeq \bm{G}$ (in the sense of positive semi-definite matrices), this regularization strengthens further the SOC constraint. Moreover, SOCP modeling frameworks (e.g.\ CVXPY or CVXGEN) suggest to replace 
	quadratic penalties (see \eqref{def_QR_loss}) with the equivalent $\sqrt{\sum_{q\in[Q]}\|f_q\|_k^2}\leq \tilde{\lambda}_f$ and 
	$ \|\bm{b}\|_2\leq \tilde{\lambda}_{\bm{b}}$ forms. This stems from their reliance on internal primal-dual interior point techniques.
	
	\section{Numerical experiments}\label{sec:numerical-demo}
	In this section we demonstrate the efficiency of the presented SOC technique to solve hard shape-constrained problems.\footnote{The code replicating our numerical experiments is available at \url{https://github.com/PCAubin/Hard-Shape-Constraints-for-Kernels}.} We focus on the task of joint quantile regression (JQR)  where the conditional quantiles are encoded by the pinball loss \eqref{def_QR_loss} and the shape requirement to fulfill is the non-crossing property \eqref{def_QR_cons}.
	Supplement (Section~\ref{sec:KRR-demo}) provides an additional illustration in kernel ridge regression (KRR, \eqref{def_KRR}) on the importance of enforcing hard shape constraints in case of increasing noise level.
	\begin{itemize}[labelindent=0cm,leftmargin=*,topsep=0cm,partopsep=0cm,parsep=0.1cm,itemsep=0cm]
		\item \tb{Experiment-1}:  We compare the performance of the proposed SOC technique on 9 UCI benchmark datasets with a state-of-the-art JQR solver relying on soft shape constraints.
		\item \tb{Experiment-2}: We augment the non-crossing constraint of JQR with monotonicity and concavity. Our two examples here belong to economics and to the analysis of aircraft trajectories.
	\end{itemize}
	
	In our experiments we used a Gaussian kernel with bandwidth $\sigma$, ridge regularization parameter $\lambda_f$ and $\lambda_{\b b}$ (or upper bounds $\tilde{\lambda}_f$ on $\sqrt{\sum_{q\in [Q]}\|f_q\|^2_k}$ and $\tilde{\lambda}_{\bm{b}}$ on $\|\bm{b}\|_2$). We learned jointly five quantile functions ($\tau_q\in\{0.1,0.3,0.5,0.7,0.9\}$). We used CVXGEN \citep{mattingley12cvxgen} to solve \eqref{opt-soc_inf}; the experiments took from seconds to a few minutes to run on an i7-CPU 16GB-RAM laptop.
	
	\begin{table}
		\caption{Joint quantile regression on 9 UCI datasets. Compared techniques: Primal-Dual Coordinate Descent (PDCD, \citealp{sangnier16joint}) and the presented SOC technique. 
			Rows: benchmarks. 2nd column: dimension ($d$). 3rd column: sample number ($N$). 4-5th columns: mean $\pm$ std of $100
			\times$value of the pinball loss for PDCD and SOC; smaller is
			better. \label{table:quantile_results}}
		\centering
		\begin{tabular}{lcrr@{\hspace{0.1cm}}c@{\hspace{0.1cm}}r r@{\hspace{0.1cm}}c@{\hspace{0.1cm}}r}
			\toprule
			Dataset & $d$ & $N$ & \multicolumn{3}{r}{PDCD} & \multicolumn{3}{c}{SOC}\\
			\midrule
			engel & $1$ & $235$ & $48$ & $\pm$ & $8$ & $53$ &$\pm$ & $9$ \\
			GAGurine & $1$ & $314$ & $61$ & $\pm$ & $7$ & $65$ & $\pm$ & $6$ \\
			geyser & $1$ & $299$ & $105$ &$\pm$ & $7$ & $108$ & $\pm$ & $3$ \\
			mcycle & $1$ & $133$ & $66$ & $\pm$ & $9$ & $62$ & $\pm$ & $5$ \\
			ftcollinssnow & $1$ & $93$ & $154$& $\pm$ & $16$ & $148$ & $\pm$ & $13$ \\			
			CobarOre & $2$ & $38$ & $159$ & $\pm$ & $24$ & $151$ & $\pm$ & $17$ \\
			topo & $2$ & $52$ & $69$ & $\pm$ & $18$ & $62$ & $\pm$ & $14$ \\
			caution & $2$ & $100$ & $88$ & $\pm$ & $17$  & $98$ &$\pm$& $22$ \\	
			ufc & $3$ & $372$ & $81$ & $\pm$ & $4$  & $87$ &$\pm$ & $6$ \\	
			\bottomrule
		\end{tabular}
	\end{table}
	In our \tb{first set of experiments} we compared the efficiency of the proposed SOC approach with the PDCD technique \citep{sangnier16joint} which minimizes the same loss \eqref{def_QR_loss} but with a \emph{soft} non-crossing encouraging regularizer. We considered $9$ UCI benchmarks. Our datasets were selected with $d\in\{1,2,3\}$; to our best knowledge none of the available JQR solvers is able to guarantee in a hard fashion the non-crossing property of the learned quantiles out of samples even in this case. Each dataset was split  into training $(70\%)$ and test $(30\%)$ sets; the split and the experiment were repeated twenty times. For each split, we optimized the hyperparameters $(\sigma,\tilde{\lambda}_f,\tilde{\lambda}_{\bm{b}})$ of SOC, searching over a grid to  minimize the pinball loss through a 5-fold cross validation on the training set. Particularly, the kernel bandwith $\sigma$ was searched over the square root of the deciles of the squared pairwise distance between the points $\{\b x_n\}_{n\in[N]}$. The upper bound $\tilde{\lambda}_f$ on $\sqrt{\sum_{q\in [Q]}\|f_q\|^2_k}$ was scanned in the log-scale interval $[-1,2]$. The upper bound $\tilde{\lambda}_{\bm{b}}$ on $\|\bm{b}\|_2$ was kept fixed:  $\tilde{\lambda}_{\bm{b}} = 10 \max_{n\in [N]}|y_n|$. We then learned a model on the whole training set and evaluated it on the test set. The covering of $K=\prod_{r \in [d]}\left[\min\{(\b x_n)_r\}_{n\in [N]},\max\{(\b x_n)_r\}_{n\in[N]}\right]$ was carried out with $\|\cdot\|_2$-balls of radius $\delta$ chosen such that the number $M$ of added points was less than $100$. This allowed for a rough covering while keeping the computation time for each run to be less than one minute. Our results 
	are summarized in Table~\ref{table:quantile_results}. The table shows that while the proposed SOC method guarantees the shape constraint in a \emph{hard} fashion, its performance is on par with the 
	state-of-the-art soft JQR solver. 
	\begin{figure}
		\centering
		\hspace*{-0.35cm}\subfloat[][]{\label{fig:QR_engel_plot_monotone}\includegraphics[height=5.9cm]{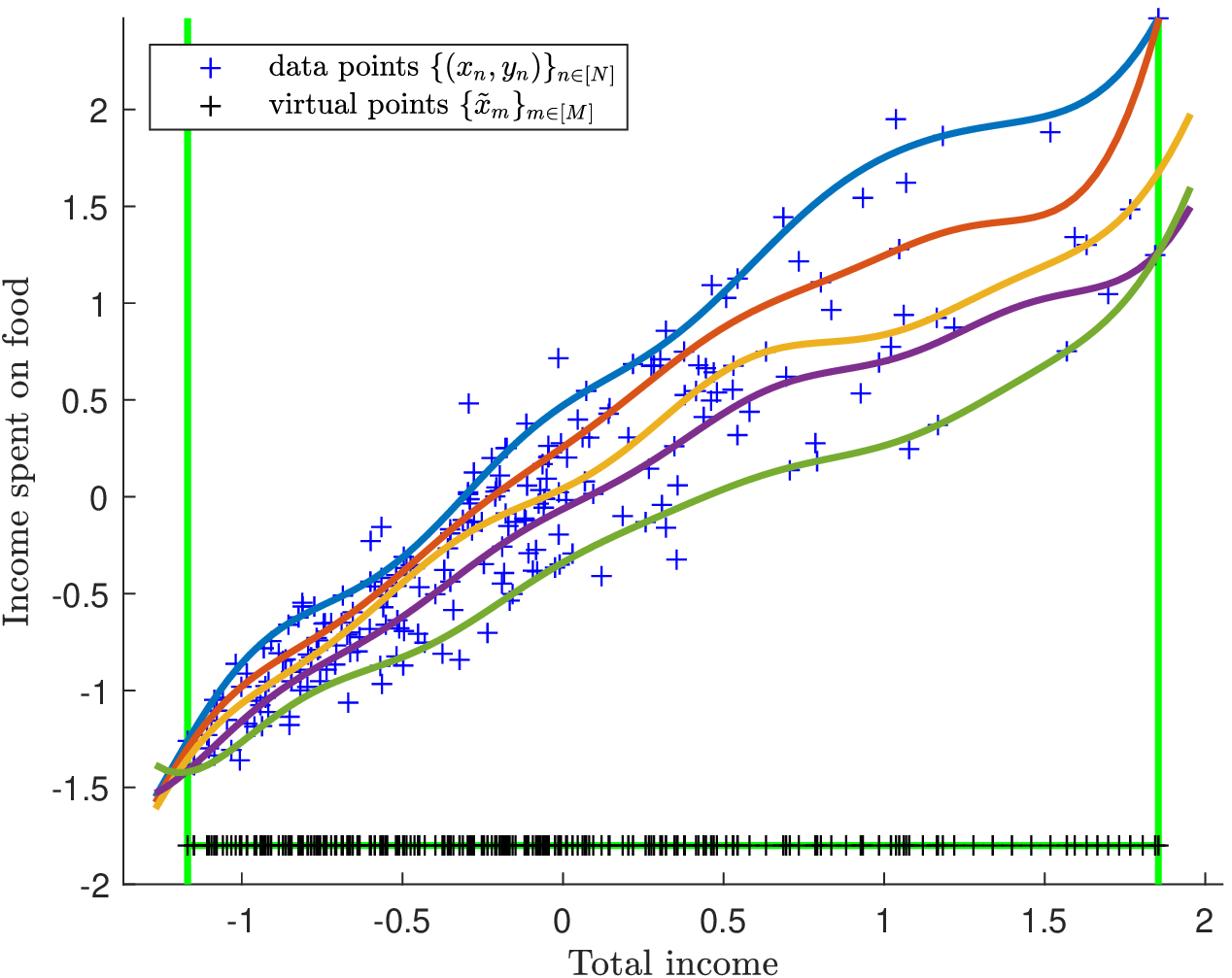}} \hspace*{-0.9cm}
		\subfloat[][]{\label{fig:QR_engel_plot_concave}\includegraphics[height=5.9cm]{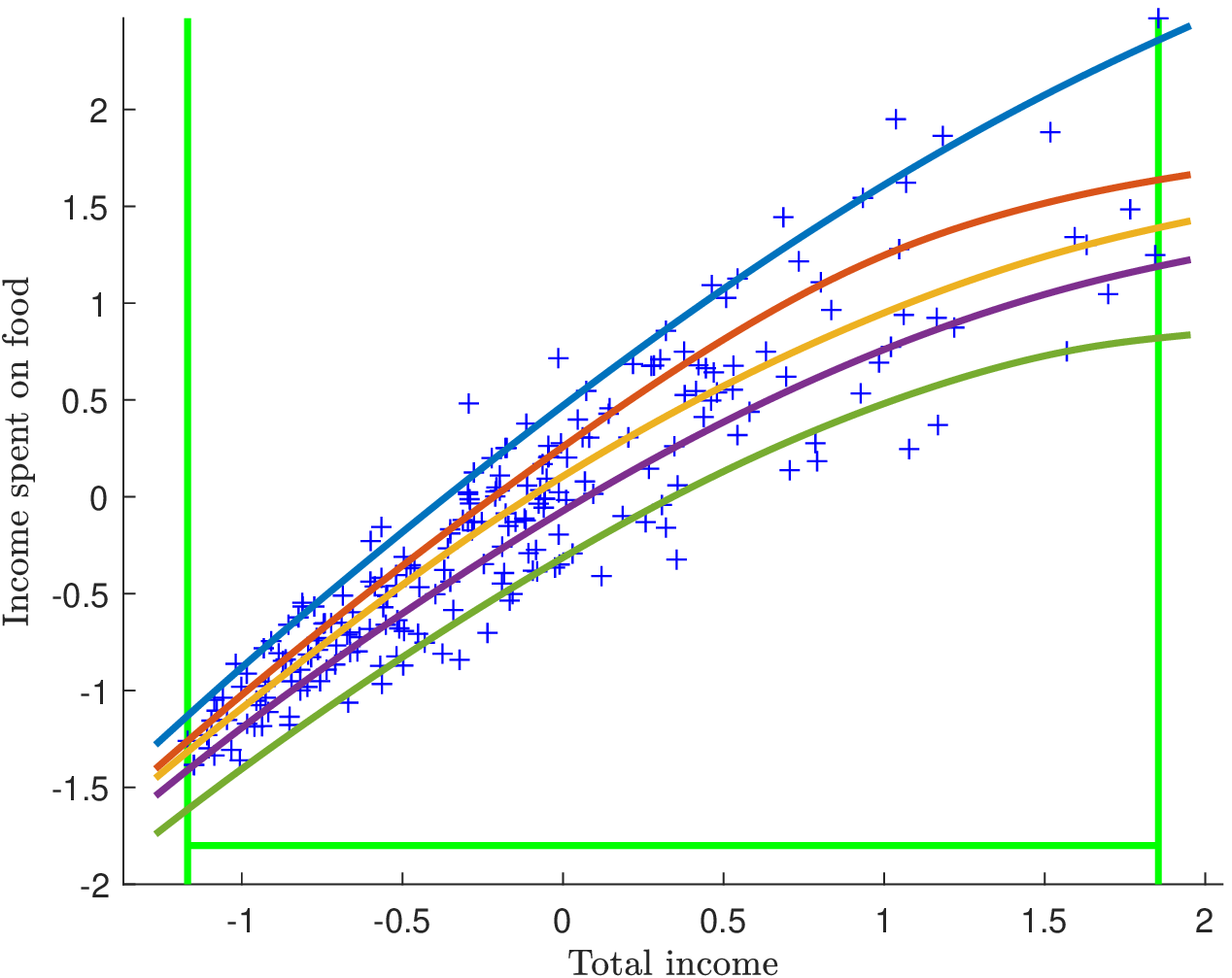}}
		\caption{Joint quantile regression on the engel dataset using the SOC technique. Solid lines: estimated conditional quantile functions with $\tau_q\in\{0.1,0.3,0.5,0.7,0.9\}$ from bottom (dark green) to top (blue). Left plot: with non-crossing \emph{and} increasing constraints. Right plot: with non-crossing, increasing  \emph{and} concavity constraints.\label{fig:QR_engel_plot}}
	\end{figure}

	In our \tb{second set of experiments}, we demonstrate the efficiency of the proposed SOC estimator on tasks with additional hard shape constraints. Our first example is drawn from \tb{economics}; we focused on JQR for the engel dataset, applying the same parameter optimization as in the first experiment. In this benchmark, the $\{(x_n,y_n)\}_{n\in[N]} \subset \R^2$ pairs correspond to annual household income ($x_n$) and food expenditure ($y_n$), preprocessed to have zero mean and unit variance. Engel's law postulates a monotone increasing property of $y$ w.r.t.\ $x$, as well as concavity. We therefore constrained the quantile functions to be non-crossing, monotonically increasing \emph{and} concave. 
	The interval $K=\left[\min\{x_n\}_{n\in [N]},\max\{x_n\}_{n\in [N]}\right]$ was covered with a non-uniform partition centered at the ordered centers $\{\tilde{x}_{m\in[M]}\}$ which included the original points $\{x_n\}_{n\in[N]}$ augmented with $15$ virtual points. 
	\begin{figure}
		\begin{center}
			\includegraphics[height=6cm]{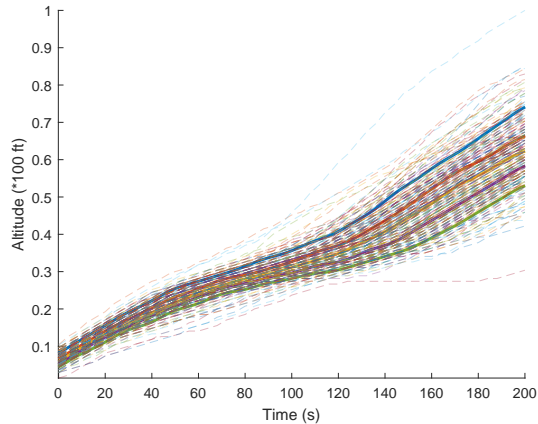}
			\caption{Joint quantile regression on aircraft takeoff trajectories. Number of samples: $N=15657$. Shape  constraints: non-crossing and increasing constraints. Dashed lines: trajectory samples. Solid lines: estimated conditional quantile functions.}\label{fig:QR_enac_plot}
		\end{center}
	\end{figure}	
	The radiuses were set to $\delta_{i,m}:=\frac{\tilde{x}_{m+1}-\tilde{x}_m}{2}$ ($m\in [M-1]$, $i\in [I]$). The estimates with or without concavity are available in Fig.~\ref{fig:QR_engel_plot}. 
	It is interesting to notice that the estimated curves can intersect outside of the interval $K$ (see Fig.~\ref{fig:QR_engel_plot}(a)), and that the additional concavity constraint mitigates this intersection (see Fig.~\ref{fig:QR_engel_plot}(b)).
	
	In our second example with extra shape constraints, we focused on the analysis of more than $300$ \tb{aircraft trajectories} \citep{nicol13functional} which describe the radar-measured altitude ($y$) of aircrafts flying between two cities (Paris and Toulouse) as a function of time ($x$). These trajectories were restricted to their takeoff phase (where the monotone increasing property should hold), giving rise to a total number of samples $N=15657$. We imposed non-crossing and monotonicity property. 
	The resulting SOC-based quantile function estimates describing the aircraft takeoffs are depicted in Fig.~\ref{fig:QR_enac_plot}. The plot illustrates how the estimated quantile functions respect the hard shape constraints and shows where the aircraft trajectories concentrate under various level of probability, defining a corridor of normal flight altitude values. 
	
	These experiments demonstrate the efficiency of the proposed SOC-based solution to hard shape-constrained kernel machines.
	
	\section{Broader impact}
	Shape constraints play a central role in economics, social sciences, biology, finance, game theory, reinforcement learning and control problems as they enable more data-efficient computation and help interpretability. The proposed principled way of imposing hard shape constraints and algorithmic solution are expected to have positive impact in the aforementioned areas.
	For instance, from social  perspective the studied quantile regression application can allow ensuring that safety regulations are better met. The improved sample efficiency, however, might result in dropping production indices and reduced privacy due to   more target-specific applications.
	
	\begin{ack}
	     ZSz benefited from the support of the Europlace Institute of Finance and that of the \href{http://www.cmap.polytechnique.fr/~stresstest/}{Chair Stress Test}, RISK Management and Financial Steering, led by the French École Polytechnique and its Foundation and sponsored by BNP Paribas.
	\end{ack}

	\bibliographystyle{apalike}
	\bibliography{./BIB/neurips_2020} 
	
	\clearpage
	\appendix
	
	\begin{center}
		{\Large\tb{Supplement}}
	\end{center}	
	
	We provide the proof (Section~\ref{sec:proofs}) of our main result  presented in Section~\ref{sec:result}. Section~\ref{sec:KRR-demo} is about an additional numerical illustration in the context of kernel ridge regression on the importance of hard shape constraints in case of increasing level of noise.  For completeness, reformulations   of  the additional shape constraint examples for \eqref{def_mixded_constraint} mentioned at the end of Section~\ref{sec:problem-formulation}  are detailed in Section~\ref{sec:shape-constraints}.

	\section{Proof}\label{sec:proofs}
	For $i\in[I]$, we shall below denote $\phi_i=(\b W \b f-\b f_{0})_i$ and $\beta_i=(\b b_0-\b U \b b)_i$. The proofs of the different parts are as follows.
	
	(i) \tb{Tightening}: By rewriting constraint \eqref{def_mixded_constraint} using the derivative-reproducing property of kernels \citep{zhou08derivative,saitoh16theory} 
	we get
	\begin{equation}
	\langle \phi_i,D_{i,\b x} k(\b x,\cdot) \rangle_k =D_i\phi_i(\b x)\geq \beta_i, \, \forall \b x\in K_i.\label{def_mixded_constraint_repro}
	\end{equation}
	Let us reformulate this constraint as an inclusion of sets
	\begin{equation*}
	\Phi_{D_i}(K_i) \subseteq H^+_{\phi_i,\beta_i}:=\{g\in \Fk\,|\, \left\langle \phi_i,g \right\rangle_k\geq \beta_i\},
	\end{equation*}
	where $\Phi_{D_i}:\b x\mapsto D_{i,\b x} k(\b x,\cdot)\in \Fk$ and $\Phi_{D_i}(X):= \{\Phi_{D_i}(\b x)\,|\, \b x\in X\}$.
	
\begin{figure}
		\centering
		\includegraphics[height=6cm]{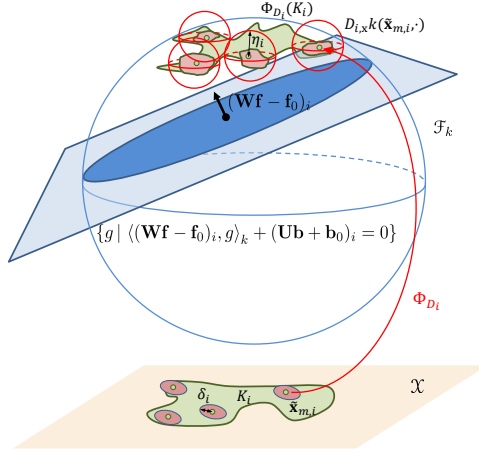}	
		\caption{Illustration of the SOC constraint \eqref{def_soc_inf}.	\label{fig:SOC_Sketch}}
	\end{figure}

	In order to get a finite geometrical description of $\Phi_{D_i}(K_i)$, we consider 
	a finite covering of the compact set $K_i$:
	\begin{equation*}
	\{\tilde{\b x}_{i,m}\}_{m\in[M_i]} \subseteq K_i \subseteq \bigcup_{m\in[M_i]} \BB_{\|\cdot\|_\X}\left(\tilde{\b x}_{i,m},\delta_{i,m}\right),
	\end{equation*}
	which implies that 
	\begin{equation*}
	\Phi_{D_i}(K_i)\subseteq \bigcup_{m\in[M_i]}\Phi_{D_i}\left(\BB_{\|\cdot\|_\X}\left(\tilde{\b x}_{i,m},\delta_{i,m}\right)\right).
	\end{equation*} 
	From the regularity of $k$, it follows that $\Phi_{D_i}$ is continuous from $\X$ to $\Fk$, and we define $\eta_{i,m}>0$ ($i\in[I]$, $m\in [M_i]$) as 
	\begin{equation}
	\eta_{i,m}:=\hspace*{-0.4cm}\sup_{\b u\in \BB_{\|\cdot\|_\X}(\b 0,1)} \|\Phi_{D_i}\left(\tilde{\b x}_{i,m}\right) - \Phi_{D_i}\left(\tilde{\b x}_{i,m}+\delta_{i,m}  \b u\right)\|_k. \label{eq:eta_{i,m}}
	\end{equation}
	This means that for all $m\in[M_i]$ 
	\begin{align*}
	\Phi_{D_i}\left(\BB_{\|\cdot\|_\X}\left(\tilde{\b x}_{i,m},\delta_{i,m}\right)\right) &\subseteq \Phi_{D_i}\left(\tilde{\b x}_{i,m}\right) + \BB_k\left(0,\eta_{i,m}\right),
	\end{align*}
	where $\BB_k(0,\eta_{i,m}) := \{g\in \Fk\,|\, \left\|g\right\|_k \le \eta_{i,m}\}$. In other words, for \eqref{def_mixded_constraint_repro} to hold, it is sufficient that
	\begin{equation}\label{incl_ball_halfspace}
	\Phi_{D_i}\left(\tilde{\b x}_{i,m}\right) + \BB_k\left(0,\eta_{i,m}\right) \subseteq H^+_{\phi_i,\beta_i},\, \forall m\in[M_i]. 
	\end{equation} 
	By the definition of $H^+_{\phi_i,\beta_i}$, \eqref{incl_ball_halfspace} is equivalent to
	\begin{align*}
	\beta_i &\leq \inf_{g\in\BB_k}  \langle \phi_i,D_{i,\b x} k(\tilde{\b x}_{i,m},\cdot) + \eta_{i,m}g\rangle_k = D_i(\phi_i)(\tilde{\b x}_{i,m}) - \eta_{i,m} \|\phi_i\|_k,\, \forall  m\in[M_i].
	\end{align*}
	Taking the minimum over $m\in[M_i]$, we get 
	\begin{equation}\label{def_soc_inf_non-unif}
	\|\phi_i\|_k \leq \min_{m\in[M_i]} \frac{1}{\eta_{i,m}}\left[-\beta_i + D_i(\phi_i)\left(\tilde{\b x}_{i,m}\right)\right].
	\end{equation}
	Hence we proved that for any $(\b f,\bm{b})$ satisfying \eqref{def_soc_inf_non-unif},  \eqref{def_mixded_constraint_repro} also holds. The SOC-based reformulation is illustrated geometrically in Fig.~\ref{fig:SOC_Sketch}. Constraint \eqref{def_mixded_constraint} can be reformulated as requiring that the image $\Phi_{D_i}(K_i)$ of $K_i$ under the $D_i$-feature map $\Phi_{D_i}(\b x):= D_{i,\b x} k(\b x,\cdot)\in \Fk$ is contained in the halfspace 'above' the affine hyperplane defined by  normal vector $(\b W \b f-\b f_{0})_i$ and bias $(\b b_0-\b U \b b)_i$. The discretization \eqref{eq:v-disc} of constraint \eqref{def_mixded_constraint} at the points $\left\{\tilde{\b x}_{i,m}\right\}_{m\in[M_i]}$ only requires the images $\Phi_{D_i}(\tilde{\b x}_{i,m})$ of the points to be above the hyperplane. Constraint \eqref{def_soc_inf} instead inflates each of those points by a radius $\eta_i$.
	
	(ii) \tb{Representer theorem}: 
	For any $q\in[Q]$, let $f_{\bm{\eta}, q}=v_q+w_q$ where $v_q$ belongs to\footnote{The linear hull of a finite set of points $(\b v_m)_{m\in[M]}$ in a vector space is denoted by $\Sp(\{\b v_m\}_{m\in[M]}) = \{\sum_{m\in [M]} a_m \b v_m \, |\, a_m \in \R,\,\, \forall m\in [M]\}$.} 
	\begin{align*}
	V & := \Sp\left(\{f_{0,i}\}_{i\in{I}},\{k(\b x_{n},\cdot)\}_{n\in[N]},\{D_{i,\b x}k(\tilde{\b x}_{i,m},\cdot)\}_{m\in[M_i], i\in[I]}\right) \subset \Fk
	\end{align*} 
	while $w_q$ is in the orthogonal complement of $V$ in $\Fk$ ($w_q \in V^{\perp}:=\{w\in \Fk: \left<w,v\right>_k=0,\, \forall v\in V\}$). Let $\b v:=(v_q)_{q\in[Q]} \in (\Fk)^Q$. As constraint \eqref{def_soc_inf} holds for $(\b f_{\bm{\eta}},\bm{b}_{\bm{\eta}})$,
	\begin{align*}
	(\b b_0-\b U \b b_{\bm{\eta}})_i + \eta_i \|(\b W \b f_{\bm{\eta}}-\b f_{0})_i\|_k
	&\le \min_{m\in[M_i]} D_i(\b W \b f_{\bm{\eta}}-\b f_{0})_i\left(\tilde{\b x}_{i,m}\right), \forall i\in [I].
	\end{align*}
	However $(\b v,\bm{b}_{\bm{\eta}})$ also satisfies \eqref{def_soc_inf} since $\|(\b W \b v-\b f_{0})_i\|_k  \le \|(\b W \b f_{\bm{\eta}}-\b f_{0})_i\|_k$ and 
	$D_i(\b W \b v-\b f_{0})_i\left(\tilde{\b x}_{i,m}\right) = D_i(\b W \b f_{\bm{\eta}}-\b f_{0})_i\left(\tilde{\b x}_{i,m}\right)$:
	\begin{align*}
	\Bigg\| \left(\b W \b f_{\bm{\eta}} - \b f_0\right)_i \Bigg\|_k^2 &= \Bigg\| \sum_{q\in [Q]} W_{i,q}\hspace*{-0.1cm}\underbrace{f_{\bm{\eta}, q}}_{v_q + w_q}  - f_{0,i} \Bigg\|_k^2 
	= \Bigg\| \underbrace{\sum_{q\in [Q]} W_{i,q} v_q  - f_{0,i}}_{\in V} + \underbrace{\sum_{q\in [Q]} W_{i,q} w_q}_{\in V^{\perp}} \Bigg\|_k^2\\
	& = \Bigg\| \sum_{q\in [Q]} W_{i,q} v_q  - f_{0,i} \Bigg\|_k^2 + \Bigg\| \sum_{q\in [Q]} W_{i,q} w_q \Bigg\|_k^2
	\ge  \Bigg\| \sum_{q\in [Q]} W_{i,q} v_q  - f_{0,i} \Bigg\|_k^2\\
	& =  \left\| (\b W \b v - \b f_0)_i \right\|_k^2,\\
	D_i(\b W \b f_{\bm{\eta}}-\b f_{0})_i\left(\tilde{\b x}_{i,m}\right) 
	&= D_i \left(\sum_{q\in [Q]} W_{i,q}  \underbrace{f_{\bm{\eta}, q}}_{v_q+w_q} -  f_{0,i} \right)\left(\tilde{\b x}_{i,m}\right)\\
	&= D_i(\b W \b v-\b f_0)_i\left(\tilde{\b x}_{i,m}\right) + D_i \left(\sum_{q\in [Q]} W_{i,q}  w_q  \right)\left(\tilde{\b x}_{i,m}\right)\\
	&= D_i(\b W \b v-\b f_0)_i\left(\tilde{\b x}_{i,m}\right) + \underbrace{\left<\sum_{q\in [Q]} W_{i,q}  w_q, D_{i,\b x} k\left(\tilde{\b x}_{i,m},\cdot\right) \right>_k}_{=0}
	\end{align*}	     
	using the derivative-reproducing property of kernels, and that $\sum_{q\in [Q]} W_{i,q}  w_q \in V^{\perp}$, while $D_{i,\b x} k\left(\tilde{\b x}_{i,m},\cdot\right) \in V$. 
	The regularizer $\Omega$ is assumed to be strictly increasing in each argument $\|f_{\bm{\eta},q}\|_k$. As $\left\|f_{\bm{\eta},q}\right\|_k^2 = \left\|v_q\right\|_k^2 + \left\|w_q\right\|_k^2$, and $(\b f_{\bm{\eta}},\b b_{\bm{\eta}})$ minimizes $\Lcal$,  $w_q=0$ necessarily; in other words $f_{\bm{\eta}, q} \in V$ for all $q\in [Q]$.
	
	(iii) \tb{Performance guarantee}: From (i), we know that the solution $(\b f_{\bm{\eta}},\b b_{\bm{\eta}})$ of \eqref{opt-soc_inf} 
	is also admissible for \eqref{opt-cons}. Discretizing the shape constraints is a relaxation of 
	\eqref{opt-cons}. Hence $v_{\text{disc}}\leq \bar{v} \leq v_{\bm \eta}$. 
	
	Let us fix any $(\b p_f,\b p_b)\in \left(\Fk\right)^Q\times\RR^P$ belonging to the subdifferential of $\Lcal(\cdot,\cdot)+\chi_{\C}(\cdot,\cdot)$ at point $(\bar{\b f},\bar{\bm{b}})$, where $\chi_\C$ 
	is the characteristic function of $\C$, i.e.\ $\chi_\C(\b u,\b v)=0$ if $(\b u, \b v)\in \C$ and $\chi_\C (\b u, \b v)=+\infty$ otherwise.
	Since $\left(\bar{\b f},\bar{\b{b}}\right)$ is the optimum of \eqref{opt-cons}, for any $(\b f,\bm{b})$ admissible for \eqref{opt-cons}, 
	\begin{align}
	\sum_{q\in[Q]}\langle p_{f,q}, f_q-\bar{f}_q \rangle_k + \langle \b p_b, \b{b}-\bar{\b{b}} \rangle_{2} &\geq 0, \label{eq:non-neg}
	\end{align}
	where $\b p_f = (p_{f,q})_{q\in [Q]}$. Hence using the $(\mu_{f_q},\mu_{\b b})$-strong convexity of $\Lcal$ w.r.t.\ $(f_q,\b b)$  we get 
	\begin{align}
	\Lcal\left(\b f_{\bm \eta},\bm{b_\eta}\right)  &\geq \Lcal\left(\bar{\b f},\bar{\bm{b}}\right) +\sum_{q\in[Q]}\left< p_{f_{\bm{\eta}},q}, f_{\bm \eta,q}-\bar{f}_q \right>_k \label{ineq_strong_convex}
	&+ \left< \b p_b, \bm{b_\eta}-\bar{\bm{b}} \right>_{2} +\sum_{q\in[Q]}\frac{\mu_{f_q}}{2} \|f_{\bm \eta,q} - \bar{f}_q\|^2_k\\
	&\quad + \frac{\mu_{\bm{b}}}{2}\left\|\bm{b_\eta}-\bar{\bm{b}}\right\|_2^2.\nonumber
	\end{align}
	As $v_{\bm \eta}-v_{\text{disc}} \geq \Lcal\left(\b f_{\bm \eta},\bm{b_\eta}\right) - \Lcal\left(\bar{\b f},\bar{\bm{b}}\right)$, using the non-negativity \eqref{eq:non-neg} for $(\b f_{\bm \eta},\bm{b_\eta})$, one gets from \eqref{ineq_strong_convex} the claimed bound 	\eqref{ineq_aposteriori}.
	
	To prove \eqref{ineq_apriori}, recall that $\left(\bar{\b f},\bar{\b{b}}\right)$ satisfies \eqref{def_mixded_constraint} and that we assume $\Bsc=\RR^P$. Let 
	$\eta_{i}=\max_{m\in[M_i]}\eta_{i,m},\,\, i\in [I]$
	with $\eta_{i,m}$ defined in \eqref{eq:eta_{i,m}}, and $\tilde{\b b}=\big(\tilde{b}_{i}\big)_{i\in [I]}\in \R^{I}$ with 
	\begin{align}
	\tilde{b}_{i} &:=\eta_i \left\|\left(\b W \bar{\b f}-\b f_{0}\right)_i\right\|_k. \label{eq:tilde-b:def}
	\end{align}
	As $\b U$ is full-row rank, one can define its right inverse ($\b U \b U^+=\b I$) as $\b U^+ =\left(\b U^{\top} \b U\right)^{-1}\b U^{\top}$. Then the pair $\left(\bar{\b f},\bar{\bm{b}}+\b U^+\tilde{\b b}\right)$ satisfies \eqref{def_soc_inf} since for any $m\in[M_i]$
	\begin{align*}
	\eta_i \left\|\left(\b W \bar{\b f}-\b f_{0}\right)_i\right\|_k &= \tilde{b}_{i} = \left(\b U \b U^+\tilde{\b b}\right)_i 
	\le \left(\b U \b U^+\tilde{\b b}\right)_i+  \underbrace{\left(\b U \bar{\b b}-\b b_0\right)_i + D_i \left(\b W \bar{\b f} - \b f_{0})_i(\tilde{\b x}_{i,m}\right)}_{\ge 0}\\
	&=  \left(\b U \left(\bar{\b b}+\b U^+\tilde{\b b}\right)-\b b_0\right)_i + D_i \left(\b W \bar{\b f} - \b f_{0})_i(\tilde{\b x}_{i,m}\right).
	\end{align*}
	Thus, $\left(\bar{\b f},\bar{\bm{b}}+\b U^+\tilde{\b b}\right)$ is admissible for \eqref{opt-soc_inf} and as 
	$(\b f_{\bm \eta},\bm{b_\eta})$ is optimal for \eqref{opt-soc_inf}, we have 
	\begin{align*}
	\Lcal\left(\b f_{\bm \eta},\bm{b_\eta}\right) - \Lcal\left(\bar{\b f},\bar{\bm{b}}\right)  &\leq  \Lcal\left(\bar{\b f},\bar{\bm{b}}+\b U^+\tilde{\b b}\right) - \Lcal\left(\bar{\b f},\bar{\bm{b}}\right)
	\stackrel{(a)}{\leq} L_b \big\|\b U^+\tilde{\b b}\big\|_2 \leq L_b\left\|\b U^+\right\| \big\|\tilde{\b b}\big\|_2 \\
	&\leq L_b\big\|\b U^+\big\| \sqrt{I}\big\|\tilde{\b b}\big\|_\infty
	\stackrel{(b)}{\le} L_b \big\|\b U^+\big\| \|\bm \eta\|_\infty \sqrt{I}\max_{i\in [I]} \left\|\left(\b W \bar{\b f}-\b f_{0}\right)_i\right\|_k \\
	&\stackrel{(c)}{=}  L_b c_f \|\bm \eta\|_\infty,
	\end{align*}
	where (a) stems from the local Lipschitz property of $\Lcal$ ($\big\|\b U^+\tilde{\b b}\big\|_2\le c_f \|\bm \eta\|_\infty$), (b) holds by \eqref{eq:tilde-b:def}, and (c) follows from the definition of 
	$c_f$.
	Combined with \eqref{ineq_strong_convex}, this gives the bound \eqref{ineq_apriori}.
	
	\section{Shape-constrained kernel ridge regression}\label{sec:KRR-demo}
	\begin{figure}
		\centering
		\hspace*{-0.2cm}\subfloat[][]{\includegraphics[height=5.7cm]{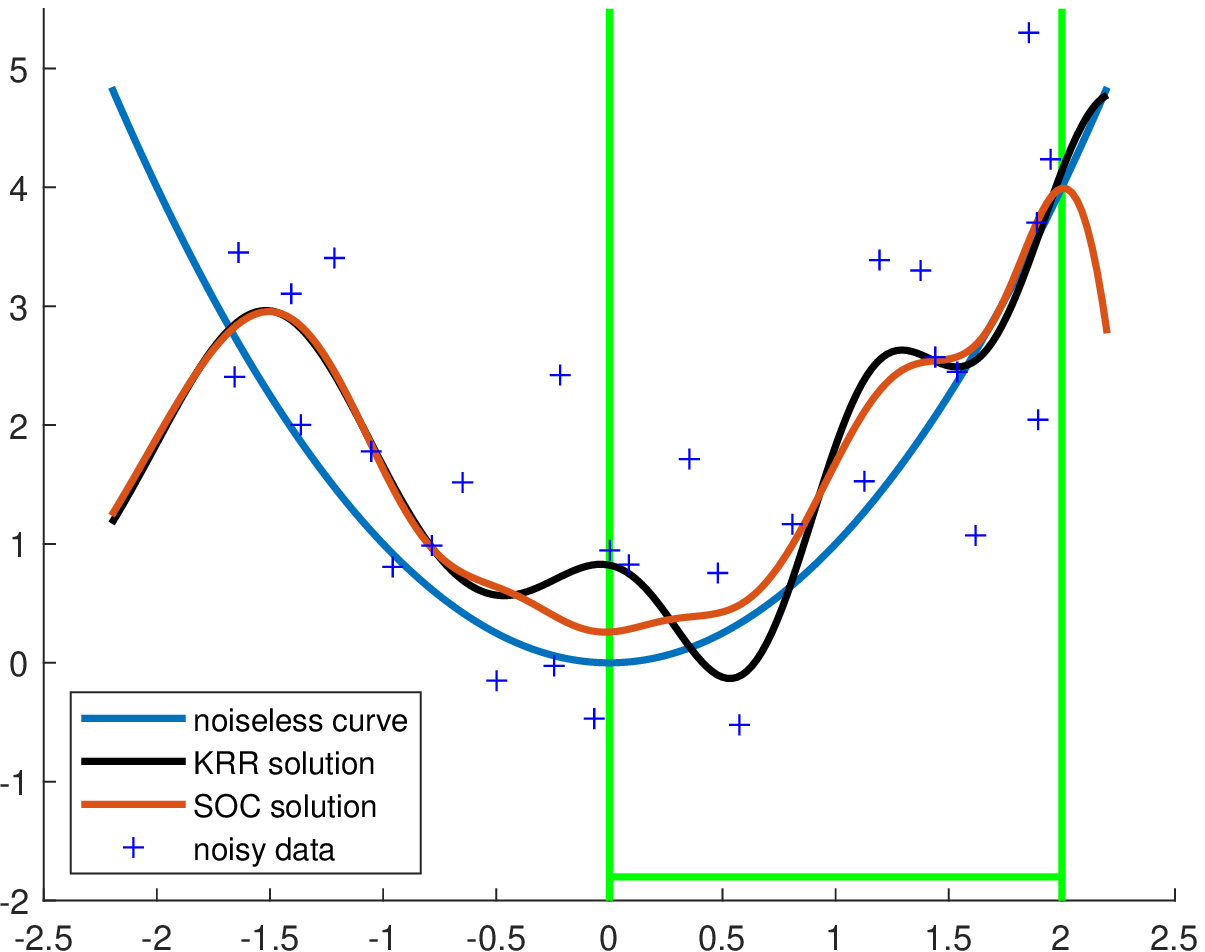}}\hspace*{-1.2cm}
		\subfloat[][]{\includegraphics[height=5.7cm]{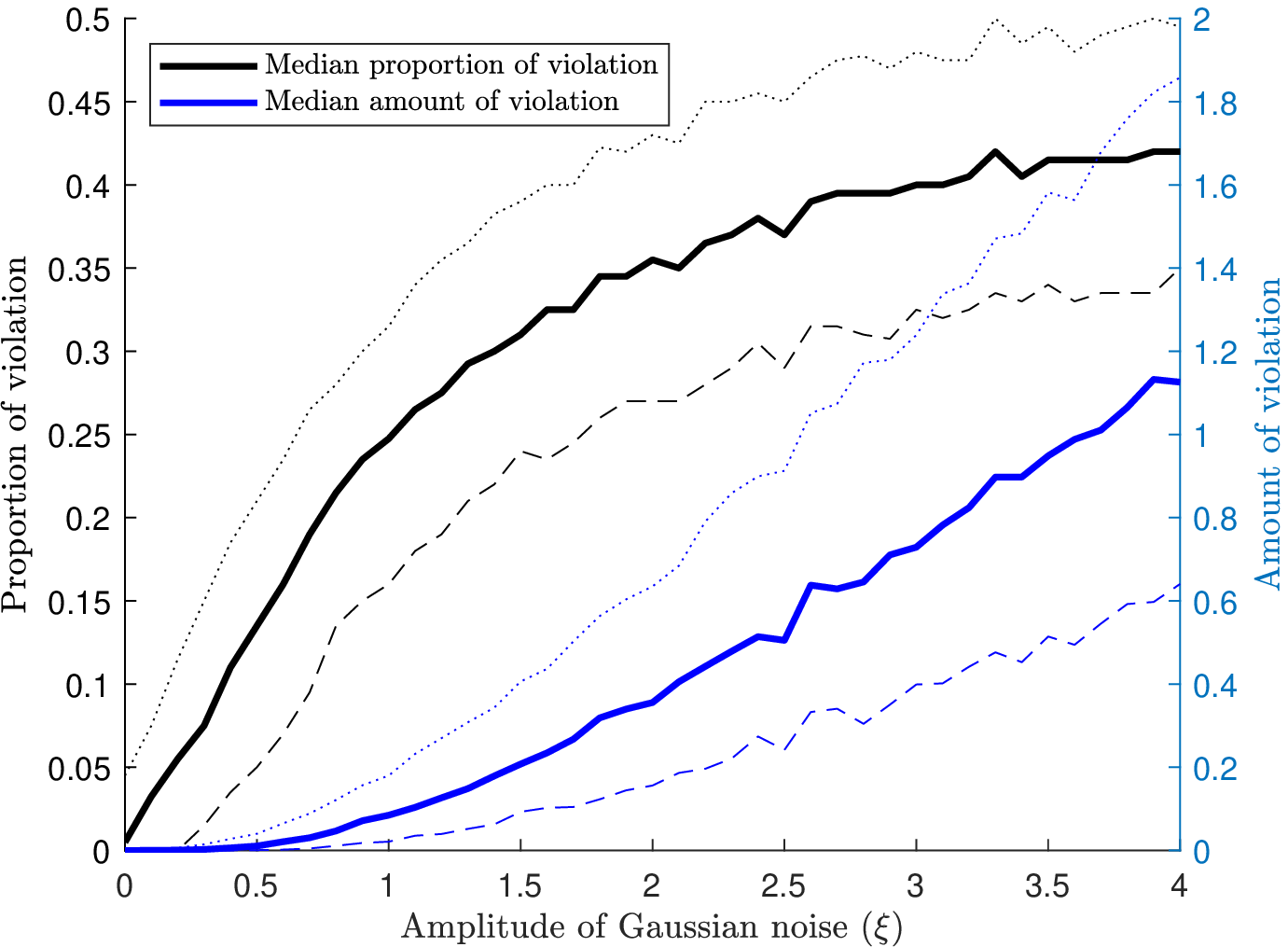}}
		
		\hspace*{-0.2cm}\subfloat[][]{\includegraphics[height=5.7cm]{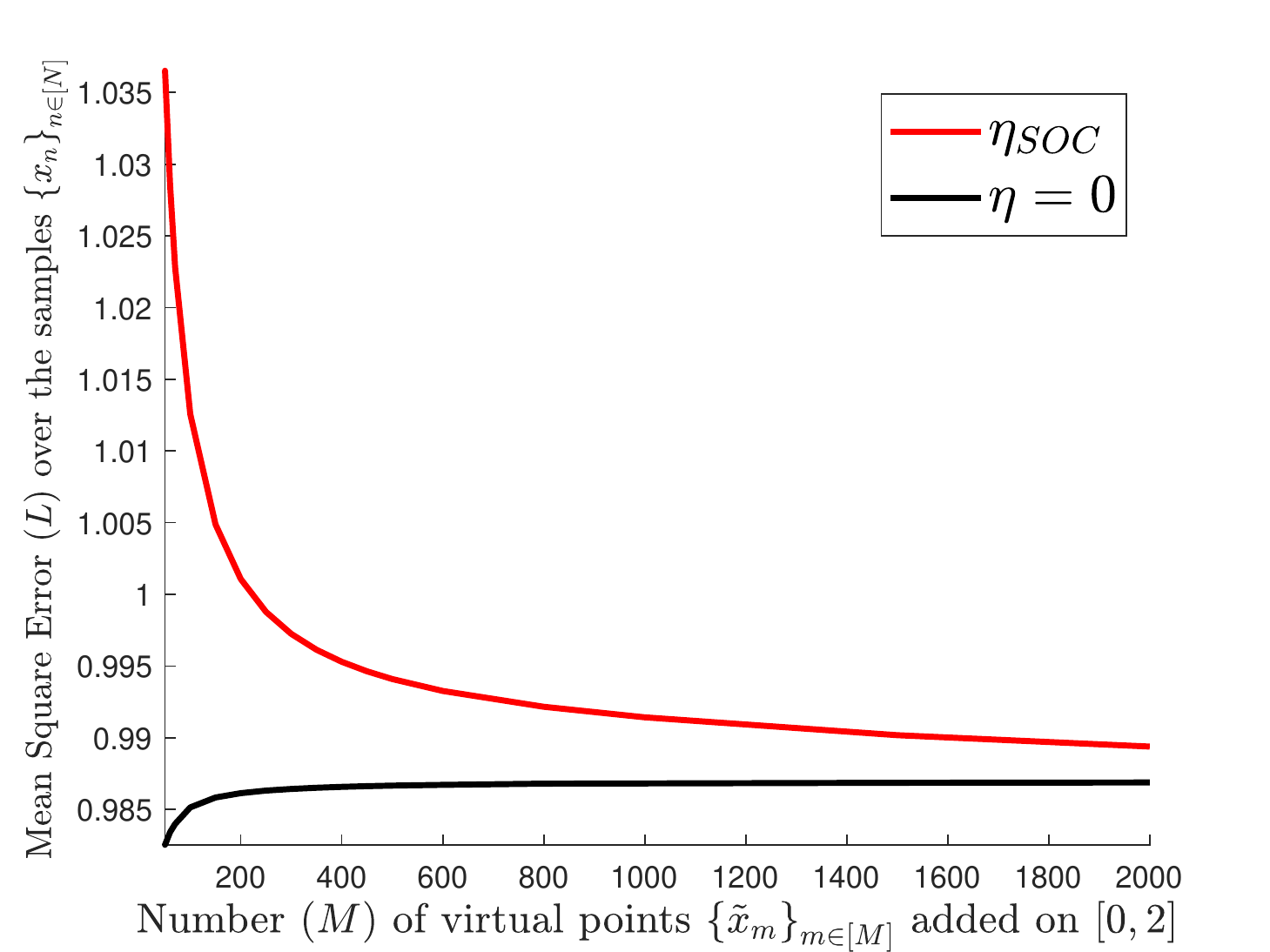}}\hspace*{-0.6cm}
		\subfloat[][]{\includegraphics[height=5.7cm]{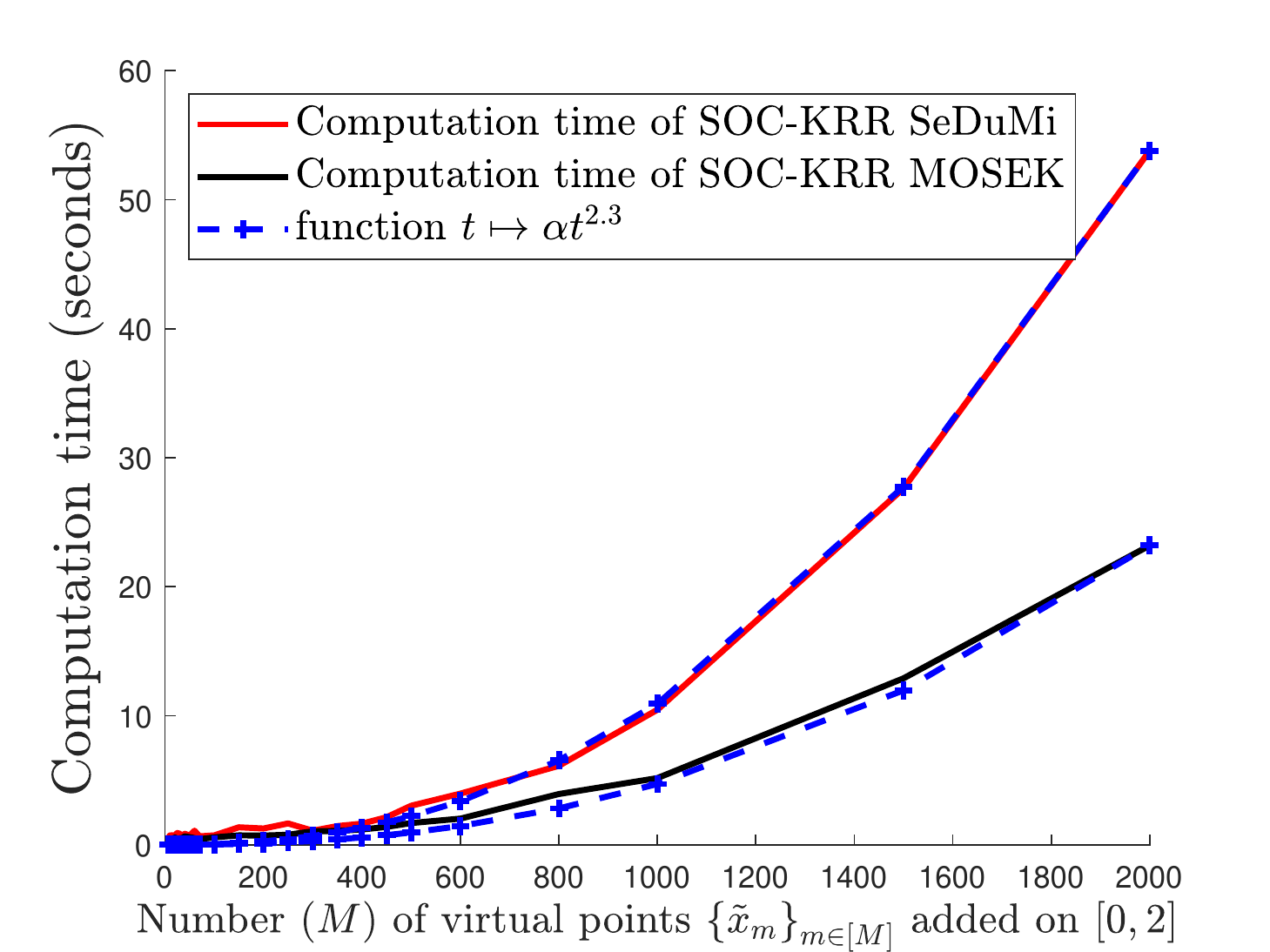}}
		\caption{(a): Illustration for kernel ridge regression. Observation: quadratic function perturbed with additive Gaussian noise. Shape constraint: monotone increasing property on $[0,2]$.  Compared techniques: regression without (KRR) and with hard shape constraint (SOC). (b): Violation of the shape constraint for the unconstrained KRR estimator as a function of the amplitude of the added Gaussian noise. Error measures: median of the proportion (left) and amount (right) of the violation of the monotone increasing property on $[0,2]$. Dashed lines: lower and upper quartiles. (c): Evolution of the optimal objective values $v_{\bm{\eta}}$ and $v_{\text{disc}}$ when increasing the number $M$ of discretization points of the constraints on $[0,2]$. (d): Computation time of \eqref{opt-soc_inf} depending on the convex optimization solver (SeDuMi or MOSEK) selected.}
		\label{fig:KRR_plot}
	\end{figure}
	In this section we illustrate in kernel ridge regression (KRR, \eqref{def_KRR}) the importance of enforcing hard shape constraints in case of increasing noise level. We consider a synthetic dataset of $N=30$ points from the graph of a quadratic function where the  values $\{x_n\}_{n\in[N]}\subset \R$ were generated uniformly on $[-2,2]$. The corresponding $y$-coordinates of the graph were  perturbed by additive Gaussian noise: 
	\begin{align*}
	y_n = x_n^2 + \epsilon_n \; (\forall n\in [N]), \{\epsilon_n\}_{n\in[N]}\stackrel{\text{i.i.d.}}{\sim}\mathcal{N}\left(0,\xi^2\right).
	\end{align*}
	We impose a monotonically increasing shape constraint on the interval $[x_l,x_u]=[0,2]$, and study the effect of the level of the added noise ($\xi$) on the desired increasing property of the estimate without (KRR) and with monotonic shape constraint (SOC). Here $\sigma=0.5$ and $\lambda_f=10^{-4}$, while $\xi$ varies in the interval $[0,4]$. 
	
	Fig.~\ref{fig:KRR_plot}(a) provides an illustration of the estimates in case of a fixed noise level $\xi=1$. There is a good match between the KRR and SOC estimates outside of the interval $[0,2]$, while the proposed SOC technique is able to correct the KRR estimate to enforce the monotonicity requirement on $[0,2]$.
	In order to assess the performance of the unconstrained KRR estimator under varying level of noise, we repeated the experiment $1000$ times
	for each noise level $\xi$ and computed the proportion and amount\footnote{These performance measures are defined as $\frac{1}{2}\int_{0}^{2} \max(0,-f'(x))\d x$ and $\int_{0}^{2} \max_{y\in[0,x]}[f(y)-f(x)]\d x$. By construction both measures are zero for SOC.} of violation of the monotonicity requirement. Our results are summarized in Fig.~\ref{fig:KRR_plot}(b). The figure shows that the error increases rapidly for KRR as a function of the noise level, and even for very low level of noise the monotonicity requirement does not hold. These experiments demonstrate that shape constraints can grossly be violated when facing noise if they are not 
	enforced in an explicit and hard fashion. To illustrate the tightening property of Theorem \ref{thm_SOC_bounds}, i.e.\ that $v_{\text{disc}}\leq \bar{v} \leq v_{\bm{\eta}}$, Fig.~\ref{fig:KRR_plot}(c) shows the evolution of the optimal values $v_{\bm{\eta}}$ and $v_{\text{disc}}$ when increasing the number of discretization points ($M$) of the constraints on the constraint interval $[0,2]$. Since by increasing $M$, we decrease $\eta$, the value $v_{\bm{\eta}}$ decreases, whereas $v_{\text{disc}}$ increases as the discretization incorporates more constraints. Larger value of $M$ naturally increases the polynomial computation time but not necessarily at the worst cubic expense, as shown in Fig.~\ref{fig:KRR_plot}(d); the choice of the solver has also importance as it may provide a factor of two gain.

	\section{Examples of handled shape constraints} \label{sec:shape-constraints}
	In order to make the paper self-contained, in this section we provide the reformulations using derivatives of the additional shape constraints briefly mentioned at the end of Section~\ref{sec:problem-formulation}:
	$n$-monotonicity ($s=n$; \citealp{chatterjee15risk}), 
	$(n-1)$-alternating monotonicity \citep{fink92kolmogorovlandau},  
	monotonicity w.r.t.\ unordered weak majorization ($s=1$; \citealp[A.7.~Theorem]{marshall11inequalities}) or w.r.t.\ product ordering ($s=1$), or supermodularity ($s=2$; \citealp[Section~2]{simchilevi14logic}).
	
	Particularly, $n$-monotonicity ($n \in \Np$) writes as $f^{(n)}(x)\ge 0$ $(\forall x)$.
	$(n-1)$-alternating monotonicity\footnote{For instance, the generator of a $d$-variate Archimedean copula can be characterized by $(d-2)$-alternating monotonicity \citep{malov01finitedimensional,mcneil09multivariate}.} ($n\in \Np$) is similar: for $n=1$ non-negativity and non-increasing properties are required; for $n\ge 2$ $(-1)^j f^{(j)}$ has to be non-negative, non-increasing and convex for $\forall j \in \{0,\ldots,n-2\}$. The other examples are
	\begin{itemize}[labelindent=0cm,leftmargin=*,topsep=0cm,partopsep=0cm,parsep=0.1cm,itemsep=0cm]
		\item \tb{Monotonicity w.r.t.\ partial ordering}: These generalized notions of monotonicity ($\b u\preccurlyeq \b v \Rightarrow f(\b u)\le f(\b v)$) rely on the partial orderings $\b u\preccurlyeq \b v$ iff $\sum_{j \in [i]}u_j \le \sum_{j \in [i]}v_j$ for all $i \in [d]$ (unordered weak majorization) and $\b u \preccurlyeq  \b v$ iff $u_i \le v_i$ $(\forall i\in [d])$ (product ordering). For $C^1$ functions mononicity w.r.t.\ the unordered weak majorization is equivalent to
		\begin{align*}
		\partial^{\b e_1}f(\b x)\ge \ldots \ge \partial^{\b e_d}f(\b x)\ge 0 \quad (\forall \b x).
		\end{align*}
		Monotonicity w.r.t.\  product ordering for $C^1$ functions can be rephrased as   
		\begin{align*}
		\p^{\b e_j}f(\b x)\ge 0,\quad  (\forall j\in [d], \quad \forall \b x).
		\end{align*}
		\item \tb{Supermodularity}: Supermodularity means that  $f(\b u\lor \b v)+f(\b u \land \b v) \ge f(\b u) + f(\b v)$ for all $\b u, \b v\in \R^d$, where maximum and minimum are meant coordinate-wise, i.e.\ $\b u\lor \b v := (\max(u_j,v_j))_{j\in [d]}$ and $\b u\land \b v := (\min(u_j,v_j))_{j\in [d]}$ for $\b u,\b v\in \R^d$.  For $C^2$ functions this property corresponds to 
		\begin{align*}
		\frac{\p^2 f(\b x)}{\p x_i \p x_j} \ge 0 \quad (\forall i\ne j \in [d], \forall \b x).
		\end{align*}
	\end{itemize}
	
\end{document}